\documentclass{article}

\PassOptionsToPackage{numbers, compress, sort}{natbib}
\usepackage[preprint]{neurips_2026}

\usepackage[utf8]{inputenc} 
\usepackage[T1]{fontenc}    
\usepackage{hyperref}       
\usepackage{url}            
\usepackage{booktabs}       
\usepackage{amsfonts}       
\usepackage{nicefrac}       
\usepackage{microtype}      
\usepackage{xcolor}         
\usepackage{amsmath}
\usepackage{xspace}
\usepackage{graphicx}
\usepackage{subcaption}
\usepackage[capitalise]{cleveref}
\usepackage{appendix}
\usepackage{wrapfig}
\usepackage{amsthm}
\usepackage{amssymb}
\usepackage{titletoc}
\usepackage{algorithm}
\usepackage{algpseudocode}
\usepackage{arydshln}
\usepackage{pifont}
\usepackage{enumitem}

\AtBeginEnvironment{appendices}{%
  \crefname{appendix}{Appendix}{Appendices}
  \crefalias{section}{appendix}
  \crefalias{subsection}{appendix}
}

\makeatletter
\DeclareRobustCommand\onedot{\futurelet\@let@token\@onedot}
\def\@onedot{\ifx\@let@token.\else.\null\fi\xspace}

\def\eg{\emph{e.g}\onedot} 
\def\ie{\emph{i.e}\onedot}

\makeatother

\title{Is the Future Compatible? Diagnosing Dynamic Consistency in World Action Models}

\author{%
  Bo-Kai Ruan \quad Teng-Fang Hsiao \quad Ling Lo \quad Hong-Han Shuai \\
  National Yang Ming Chiao Tung University \\
  \texttt{\{bkruan.ee11,tfhsiao.ee13,linglo.ee08,hhshuai\}@nycu.edu.tw} \\
}

\begin{document}

\maketitle

\begin{abstract}

World Action Models (WAMs) enable decision-making through imagined rollouts by predicting future observations and actions. However, the reliability of these imagined futures remains under-examined: \textit{is a generated future merely visually plausible, or is it dynamically compatible with the action sequence it claims to model?} In this work, we identify \emph{action-state consistency}, the alignment between predicted actions and induced state transitions, as a missing reliability axis for WAMs. Through a systematic study across representative joint-prediction and inverse-dynamics models, we find that action-state consistency systematically separates successful and failed rollouts across many tasks and follows similar success-failure trends as learned value estimates. These results suggest that consistency captures decision-relevant structure beyond visual realism. We further identify \emph{background collapse} as an important boundary condition, where low-dynamics failed trajectories can become deceptively consistent because static futures are easier to predict. Building on these findings, we introduce a value-free consensus strategy for test-time selection, which ranks candidate rollouts by agreement among predicted futures. This strategy improves success rates on RoboCasa and RoboTwin 2.0 without additional training or reward modeling. Taken together, our findings establish action-state consistency as both a diagnostic tool for evaluating WAM reliability and a practical signal for value-free planning.

\end{abstract}
\section{Introduction}

Vision-language-action (VLA) models have recently emerged as a scalable paradigm for robot control, mapping visual observations and language instructions directly to executable actions~\citep{pi0,pi05,openvla,rt2}. By leveraging large-scale vision-language pretraining and robot demonstration data, these models provide a general interface for instruction-conditioned manipulation across diverse tasks. However, direct action prediction alone does not explicitly represent the future consequences of an action. A policy may output a plausible command without revealing whether the resulting physical transition is feasible or aligned with the intended task outcome. World Action Models (WAMs) have been proposed to address this limitation by extending VLA models with future prediction~\citep{motus,worldvla,cosmospolicy,lingbotva,dreamzero,gigaworld,fastwam}. While conventional VLA policies directly predict the next action, WAMs additionally model the future state induced by an action. This joint modeling gives WAMs a mechanism to anticipate object motion, contact, scene evolution, and the consequences of robot behavior before or during execution.

Nevertheless, the ability to generate plausible future observations does not guarantee that these predictions are action-conditioned, physically faithful, or decision-relevant. For a WAM to provide a reliable basis for decision making, the future it predicts should be consistent with the action it conditions on, whether that action is predicted by the policy, evaluated as a candidate, or executed in the environment. In other words, rather than merely generating visually plausible futures, a WAM should predict futures that are consistent with the given action and the underlying environment dynamics. We refer to this requirement as action-state consistency, a key reliability property for action-conditioned world modeling. When the transition observed after executing an action agrees with the future predicted under that action, it provides evidence that the model captures action-conditioned dynamics. In contrast, a mismatch suggests a potential failure of action conditioning: the model may produce visually plausible but dynamically incompatible futures, misrepresent contact or object motion, or fail to faithfully model the consequences of its selected actions.

Despite the central role of action-conditioned prediction in WAMs, the consistency between actions and their predicted or realized future states remains under-examined as a reliability property. Existing WAM and VLA studies have made substantial progress in policy learning, action-conditioned future prediction, verifier-based selection, reward estimation, and value-based planning~\citep{motus,worldvla,cosmospolicy,lingbotva,dreamzero,gigaworld,fastwam,rover,evolvevla,scalingverification,vlareasoner}. However, these evaluations often assess downstream success, predicted reward, or generated future quality without isolating whether the predicted future is faithful to the action-conditioned transition being evaluated. We argue that action-state consistency is a missing reliability axis for WAMs: it evaluates whether a model's predicted future is not only visually plausible, but also dynamically compatible with the action it claims to model.

In this work, we move beyond terminal reward metrics to investigate the internal reliability of WAMs through the lens of action-state consistency. We conduct our study across two representative WAM formulations: \textit{joint-prediction} models, which generate future observations and actions jointly, and \textit{inverse-dynamics} models, which infer actions conditioned on predicted future states. Our contributions are three-fold:
\begin{itemize}[leftmargin=*]
    \item \textbf{Characterizing action-state consistency in WAMs.} We show that action-state consistency is measurable across both joint-prediction and inverse-dynamics WAMs, and that it is strongly associated with task success. A simple classifier using normalized episode-level consistency scores predicts successful versus failed rollouts with AUCs of 0.77 and 0.88.

    \item \textbf{Identifying when consistency becomes unreliable.} We uncover a failure mode, which we call \textit{background collapse}, where low-dynamics failed trajectories remain visually static or collapse toward background-like predictions, making them easy to predict and therefore deceptively consistent. We show that this mismatch is explained by temporal latent transition magnitude, providing a diagnostic signal for when consistency becomes less reliable.

    \item \textbf{Using consistency as a value-free test-time selection signal.} We show that consistency follows similar success-failure trends as learned value predictions, suggesting that it can serve as a value-free ranking signal when no explicit value head is available. Consistency-guided selection improves average success rate from 66.6\% to 67.3\% on \textbf{RoboCasa} and from 90.2\% to 93.0\% on \textbf{RoboTwin 2.0}, without additional reward modeling or training.
\end{itemize}

Together, these results establish action-state consistency as both a reliability diagnostic for WAMs and a model-agnostic signal for guiding test-time decision making.

\section{World Action Models and Action-State Consistency}

WAMs aim to model both observations and actions in sequential decision-making. Let $t$ denote the current timestep, $o_{0:t}$ the past of visual observations, $q_t$ the proprioceptive state, and $\mathcal{G}$ the task specification (\eg, a language instruction). The objective is to predict a future trajectory over a horizon $\Delta$ consisting of future observations $o_{t+1:t+\Delta}$ and actions $a_{t+1:t+\Delta}$.

\paragraph{Joint-Prediction WAMs.}
In the joint prediction formulation, adopted by models such as Cosmos-Policy~\cite{cosmospolicy} and FastWAM~\cite{fastwam}, the model directly parameterizes the full joint distribution:
\begin{equation}
p(o_{t+1:t+\Delta}, a_{t+1:t+\Delta} \mid o_{0:t}, q_t, \mathcal{G}).
\end{equation}
Future observations and actions are generated simultaneously within a unified model. This formulation can be interpreted as learning an implicit world simulator together with a policy. Consequently, the compatibility between actions and future observations is enforced implicitly through a shared latent representation, making consistency an intrinsic property of the model distribution.

\paragraph{Inverse-Dynamics WAMs.}
Another formulation makes the joint distribution into two components:
\begin{equation}
p(o_{t+1:t+\Delta}, a_{t+1:t+\Delta} \mid o_{0:t}, q_t, \mathcal{G})
=
\underbrace{p(o_{t+1:t+\Delta} \mid o_{0:t}, q_t, \mathcal{G})}_{\text{world model}}
\;
\underbrace{p(a_{t+1:t+\Delta} \mid o_{0:t+\Delta}, q_t)}_{\text{inverse dynamics}}.
\end{equation}
This decomposition separates the prediction of future observations from action inference. Such a formulation is closely related to approaches such as LingBot-VA~\cite{lingbotva} and GigaWorld~\cite{gigaworld}, where the model first predicts or samples future states and subsequently infers the actions required to realize them. In this view, actions are not jointly generated with futures but are derived from them, making the alignment between action and outcome an explicit modeling objective. A hybrid variant is adopted in models such as DreamZero~\cite{dreamzero}, which perform joint prediction during training while implicitly relying on an inverse dynamics mechanism at inference time.

For simplicity, we adopt \textbf{Cosmos-Policy} pretrained on \textbf{RoboCasa}~\cite{robocasa} and \textbf{LingBot-VA} pretrained on \textbf{RoboTwin 2.0}~\cite{robotwin} as representative instances of the joint prediction and inverse dynamics formulations, respectively, for further study throughout the paper.

\paragraph{Measuring Action-State Consistency.}
We define action-state consistency as a scalar measure of how well the predicted future observations match the real observations obtained by executing the predicted actions in the environment. 
For each trajectory timestep $t$, we compute a consistency score that measures the similarity between the future observation and the real observation. Formally, given an observation $o_t$, we define the consistency score $c_t$ at horizon $\Delta$ as:
\begin{equation}\label{eq:con_score}
c_t(o_{t+\Delta}, \hat{o}_{t+\Delta}) = \exp\big( -\alpha \cdot d(o_{t+\Delta}, \hat{o}_{t+\Delta})\big),
\end{equation}
where $\hat{o}_{t+\Delta}$ denotes the predicted observation, $\alpha$ is a scaling term\footnote{We use $\alpha=0.1$ by default. The specific value does not affect our conclusions, since consistency is only used as a relative ranking signal rather than an absolute calibrated score.}, and $d(\cdot, \cdot)$ is a distance function. 

We adopt MSE as the distance metric computed in the latent space prior to VAE decoding, which avoids sensitivity to pixel-level variations~\cite{revrl,worldinworld} and ensures that the action-state consistency reflects structural agreement in the decision-relevant feature space.
\section{Characterizing Action-State Consistency}\label{sec:study}

For a WAM to serve as a reliable basis for decision making, its predicted futures should remain faithful to the action-conditioned state transitions they are intended to represent. While prior work has primarily focused on improving predictive performance or downstream task success, the dynamic alignment between imagined futures and realized environment transitions remains under-examined. We study action-state consistency as a reliability criterion along three axes:
\begin{enumerate}[leftmargin=*]
\item \textbf{Emergence and Separability (\ref{sec:rq1})}: We examine whether consistency systematically separates successful and failed rollouts across tasks and models.
\item \textbf{Boundary Conditions (\ref{sec:rq2})}: We identify when consistency becomes a less reliable indicator of task success, with a focus on the ``background collapse'' regime.
\item \textbf{Alignment with Utility (\ref{sec:rq3})}: We examine whether consistency follows similar trends as learned value predictions, assessing its potential as a value-free signal for decision making.
\end{enumerate}

\subsection{Emergence and Separability}\label{sec:rq1}

We first examine whether action-state consistency is systematically associated with task outcomes across different WAM formulations and task settings. This leads to our first research question:

\textbf{RQ1: \textit{Do WAMs exhibit measurable action-state consistency, and how does this consistency vary across models, tasks, and outcomes?}}

To answer this, we analyze \textbf{(1)} whether consistency is systematically related to task success, and \textbf{(2)} whether this relationship holds across diverse task settings. To enable comparisons across tasks with different consistency score ranges, we normalize consistency scores within each task. Specifically, for each task, we compute the mean and standard deviation of episode-level consistency and convert each episode into a z-score. This yields a task-independent measure that reflects how consistent an episode is relative to others within the same task. We then pool normalized scores across all tasks and compare their distributions for successful and failed trajectories.

\begin{figure}[h]
    \centering
    
    \begin{subfigure}{0.49\linewidth}
        \centering
        \textbf{{\small \textcolor[HTML]{212121}{(a) Cosmos-Policy}}}
        \includegraphics[width=\linewidth]{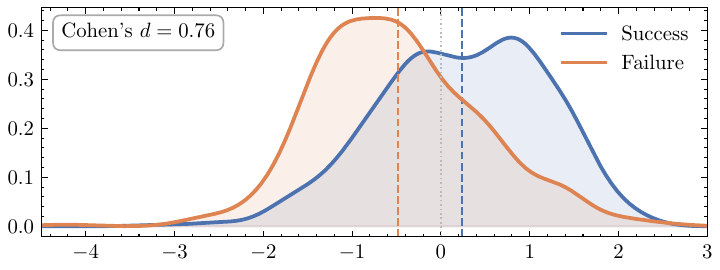}
    \end{subfigure}
    \hfill
    \begin{subfigure}{0.49\linewidth}
        \centering
        \textbf{{\small \textcolor[HTML]{212121}{(b) LingBot-VA}}}
        \includegraphics[width=\linewidth]{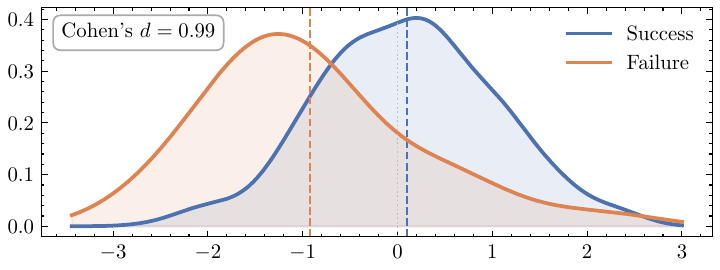}
    \end{subfigure}
    \caption{\textbf{KDE plot of consistency vs. task outcome.}
The $x$-axis shows normalized consistency ($z$-score), and the $y$-axis shows density. Across both models, \textcolor[HTML]{4C72B0}{successful trajectories} exhibit higher relative consistency compared to \textcolor[HTML]{DD8452}{failed ones}, as indicated by the effect sizes (Cohen’s $d$).}
    \vspace{-15pt}
    \label{fig:zscore_all_tasks}
\end{figure}

\begin{wrapfigure}[20]{r}{0.39\textwidth}
    \centering
    \includegraphics[width=\linewidth]{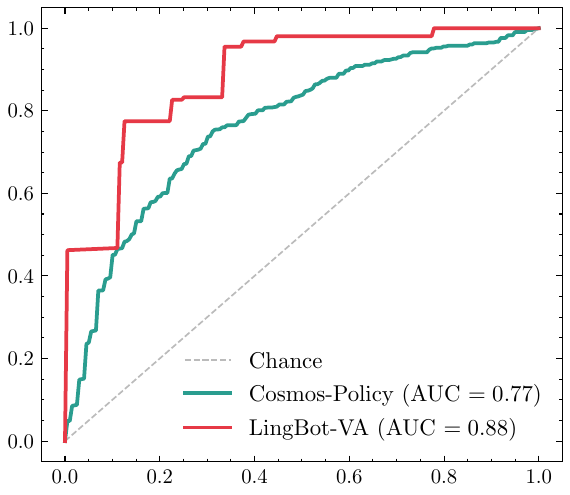}
    \vspace{-15pt}
    \caption{\textbf{ROC curves for success and failure prediction in aligned task settings.} The $x$-axis denotes the false positive rate, and the $y$-axis denotes the true positive rate. The results show that consistency can provide a predictive signal for task success in aligned task settings.}\label{fig:roc_curve}
\end{wrapfigure}

As shown in~\cref{fig:zscore_all_tasks}, successful episodes are shifted toward higher normalized consistency scores, whereas failed episodes tend to receive lower scores. This trend holds across both joint prediction and inverse dynamics formulations. Quantitatively, the two groups are statistically well-separated: on the joint prediction setting we observe a medium-to-large effect size and on the inverse dynamics setting a large effect size with Cohen's $d$ of 0.76 and 0.99, respectively, both indicating that successful and failed episodes occupy meaningfully distinct regions of the normalized consistency-score distribution. We further provide per-task consistency scores in~\cref{sec:consistency_across_tasks}, which show similar trends. Together, these results suggest that WAMs produce internally consistent predictions, and that this consistency is meaningfully related to task success.

To further assess whether consistency provides a predictive signal of task outcomes, we train a simple logistic regression classifier using normalized consistency scores and perform 5-fold cross-validation to predict success or failure. We report this classifier analysis on the consistency-aligned tasks, where successful episodes have higher average consistency than failed episodes.\footnote{Full per-task results, including weaker or reversed cases, are provided in~\cref{sec:auc_across_task}.} This ensures that the evaluated signal reflects a consistent relationship between prediction accuracy and task success. As shown in~\cref{fig:roc_curve}, consistency alone provides a strong predictive signal, achieving an AUC of 0.77 for Cosmos-Policy and 0.88 for LingBot-VA. This indicates that episodes with higher consistency scores are more likely to succeed, and that the separation observed in Figure~\ref{fig:zscore_all_tasks} translates into meaningful predictive performance. These results demonstrate that, when aligned with task outcomes, consistency serves as an effective predictor of task performance. We also provide per-task results and analysis in~\cref{sec:auc_across_task}.

\subsection{Boundary Conditions}\label{sec:rq2}

While the previous section shows that action-state consistency generally separates successful and failed episodes, we also observe several tasks in~\cref{sec:consistency_across_tasks} where this trend is reversed. In these tasks (\eg, \texttt{TurnOnSinkFaucet}), failed episodes can unexpectedly obtain higher consistency scores than successful ones. This motivates the following question:

\textbf{RQ2: \textit{Under what conditions does action-state consistency fail to predict task success, and what factors explain this mismatch?}}

\begin{figure}[h]
    \centering
    \vspace{-10pt}
    \includegraphics[width=\linewidth]{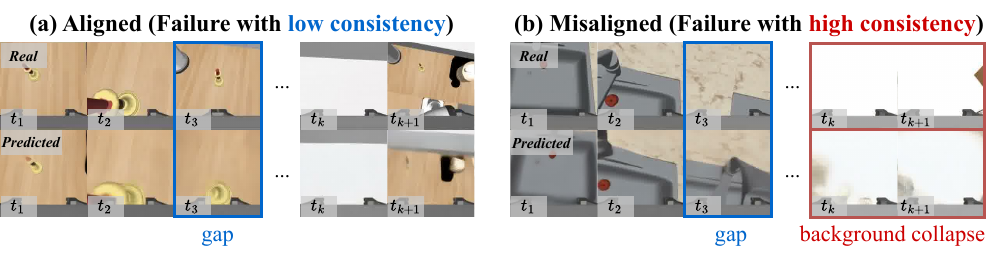}
    \vspace{-15pt}
    \caption{\textbf{Illustration of background collapse.} We compare two failed trajectories. In \textbf{(a)}, a consistency-aligned failure case with visible actions and continued scene changes. In \textbf{(b)}, a consistency-misaligned failure case, the predicted trajectory collapses toward a static background with minimal visible scene change, which can yield favorable consistency scores despite task failure.}
    \vspace{-10pt}
    \label{fig:case_illustration}
\end{figure}

To study this phenomenon, we group tasks by whether consistency separates successful and failed episodes. A task is \textbf{consistency-aligned} if successful episodes obtain higher consistency scores than failures, and \textbf{consistency-misaligned} otherwise. Inspection of misaligned tasks reveals a common failure pattern, shown in~\cref{fig:case_illustration}: failed trajectories often exhibit a phenomenon we term \textbf{background collapse}, where the predicted scene becomes nearly static, object motion weakens, and actions make little visible progress. Although these episodes fail, their static predictions are trivially accurate, yielding deceptively high consistency scores. More examples are provided in~\cref{sec:add_bg_cases}.

To validate this hypothesis, we quantify temporal scene change by the latent change magnitude. Let $z_t$ be the latent representation at the trajectory state $t$; we define the latent change $\Delta z_t$:
\begin{equation}
\Delta z_t = d(z_t, z_{t+\Delta}). 
\end{equation}
Larger $\Delta z_t$ indicates stronger motion, while smaller $\Delta z_t$ corresponds to a nearly static scene. We show the relationship in~\cref{sec:relationship} that the two quantities exhibit a negative correlation: smaller $\Delta z_t$ tend to receive higher consistency scores and vice versa. Additionally, the two signals often evolve in opposite directions over time: when $\Delta z_t$ rises, indicating stronger motion or larger state change, the consistency score typically falls; when $\Delta z_t$ decreases, consistency tends to increase.

\begin{figure}[h]
    \vspace{-5pt}
    \centering
    \begin{subfigure}{0.49\linewidth}
        \centering
        \textbf{{\small \textcolor[HTML]{212121}{(a) Cosmos-Policy}}}
        \includegraphics[width=\linewidth]{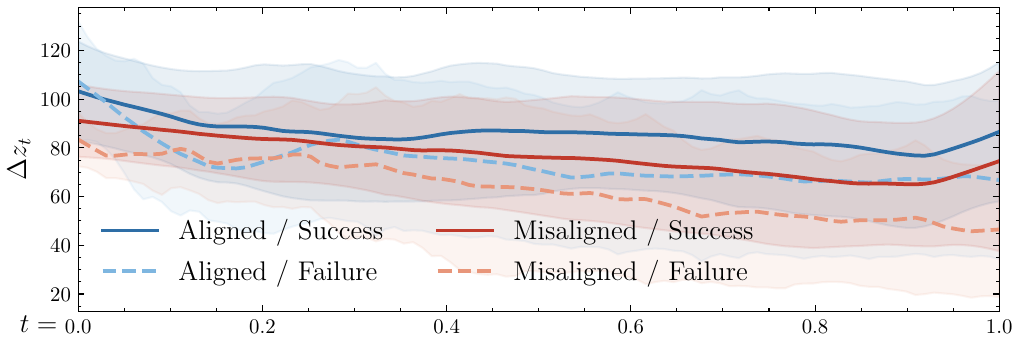}
    \end{subfigure}
    \hfill
    \begin{subfigure}{0.49\linewidth}
        \centering
        \textbf{{\small \textcolor[HTML]{212121}{(b) LingBot-VA}}}
        \includegraphics[width=\linewidth]{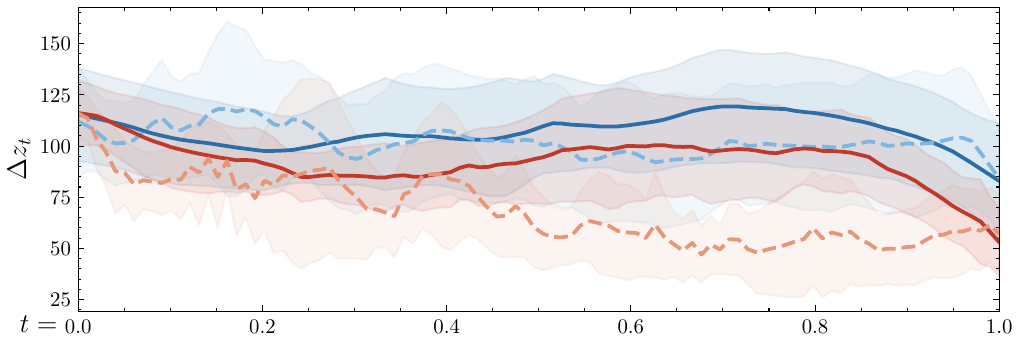}
    \end{subfigure}
    \caption{\textbf{Decomposition of latent change across success and failure episodes.} We compare latent change ($\Delta z_t$) trajectories across \textcolor[HTML]{2E6EA6}{Aligned} and \textcolor[HTML]{C0392B}{Misaligned} tasks, further partitioned into Success and Failure episodes. Notably, \textcolor[HTML]{E8967A}{Misaligned / Failure} exhibits a rapid and severe drop in latent change over time. This supports the background-collapse explanation, where extremely low $\Delta z_t$ cases can yield deceptively high consistency.}
    \label{fig:latent_consistency}
\end{figure}

This result helps explain misaligned tasks. In low-dynamics regimes, prediction becomes easier, so trivial forecasts, such as static backgrounds, can obtain high consistency scores despite not solving the task. In contrast, meaningful object interaction or rapid motion is harder to predict and may receive lower consistency scores even when behaviorally successful. This is further supported by~\cref{fig:latent_consistency}, where misaligned tasks show smaller latent changes on average, suggesting that they are more dominated by low-motion or collapsed dynamics.

Overall, these findings do not imply that action-state consistency breaks down as a useful signal. Rather, they show that consistency can be confounded by scene transition difficulty in a specific failure regime: once the model enters a stalled state, the later predicted trajectory may collapse toward a nearly static background, making failed rollouts artificially easy to predict. Since background collapse is observed mainly in \textit{later steps of misaligned failures}, the policy can be seen as entering an unrecoverable regime early on. A natural mitigation is therefore to strengthen action selection at initial trajectory steps, reducing the likelihood of such failures propagating into collapsed predictions. In~\cref{sec:mitigate_bgc}, we show that stronger early-stage test-time selection indeed improves consistency and reduces the occurrence of background-collapse failures.

\subsection{Alignment with Utility }\label{sec:rq3}

After characterizing the properties of action-state consistency, we next investigate whether this signal can be connected to downstream outcomes. In many sequential decision-making tasks, reward is sparse and often only revealed at the terminal step, making it difficult to directly assess the quality of intermediate decisions. The value function addresses this challenge by estimating expected future return before the final outcome is observed. This naturally raises the question of whether consistency also carries information about future utility, and whether it can provide a value-free decision signal when an explicit value predictor is unavailable. Concretely, we study:

\textbf{RQ3: \textit{Can action-state consistency serve as a value-free signal for downstream decision making, and how does it complement reward or value prediction?}}

\begin{wrapfigure}[15]{r}{0.4\textwidth}
    \vspace{-\baselineskip}
    \centering
    \includegraphics[width=\linewidth]{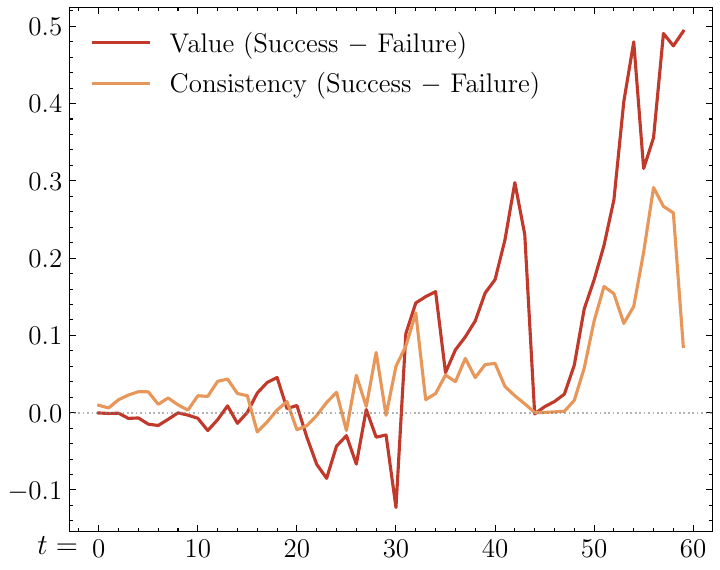}
    \caption{\textbf{Comparison of value prediction and consistency over time.}}
\label{fig:value_consistency}
\end{wrapfigure}

To examine this question, we use \textbf{Cosmos-Policy} as the reference model, since it explicitly exposes value prediction outputs. We run different random seeds to obtain a distribution of rollouts, and then compare successful and failed trajectories of the same task by measuring the gap between them. Specifically, in~\cref{fig:value_consistency}, we report the difference between successful and failed rollouts, \ie, success minus failure, for both value prediction and consistency scores. 
The positive gap indicates that successful trajectories generally obtain larger predicted values and higher consistency scores. This separation is meaningful in two ways. First, the higher predicted values ($>0$) align with their larger discounted Monte Carlo returns, confirming that the value predictor is well-calibrated. Second, the similar positive gap observed for consistency suggests that it follows the same success-failure trend as value prediction. This finding provides a direct answer to RQ3: beyond measuring model reliability, consistency also captures information correlated with downstream utility. In practice, this suggests that consistency can serve as a decision-making signal when a WAM does not explicitly provide a value head.

\section{Consistency-Guided Test-Time Selection}\label{sec:consistency_guided}

Many WAMs predict future observations and actions, but do not necessarily provide an explicit value prediction head. In such cases, selecting among candidate futures remains challenging because task rewards are sparse and typically observed only at episode termination. The diagnostic form of action-state consistency compares predicted futures with realized transitions, but directly using this signal for test-time selection would require executing each candidate branch. To obtain a deployable alternative, we also consider an internal proxy based on agreement among sampled future predictions. We compare value-based selection with two consistency-guided best-of-$N$ strategies: environment-exploring selection and future-consensus selection.

\begin{figure}[h]
    \centering
    \includegraphics[width=\linewidth]{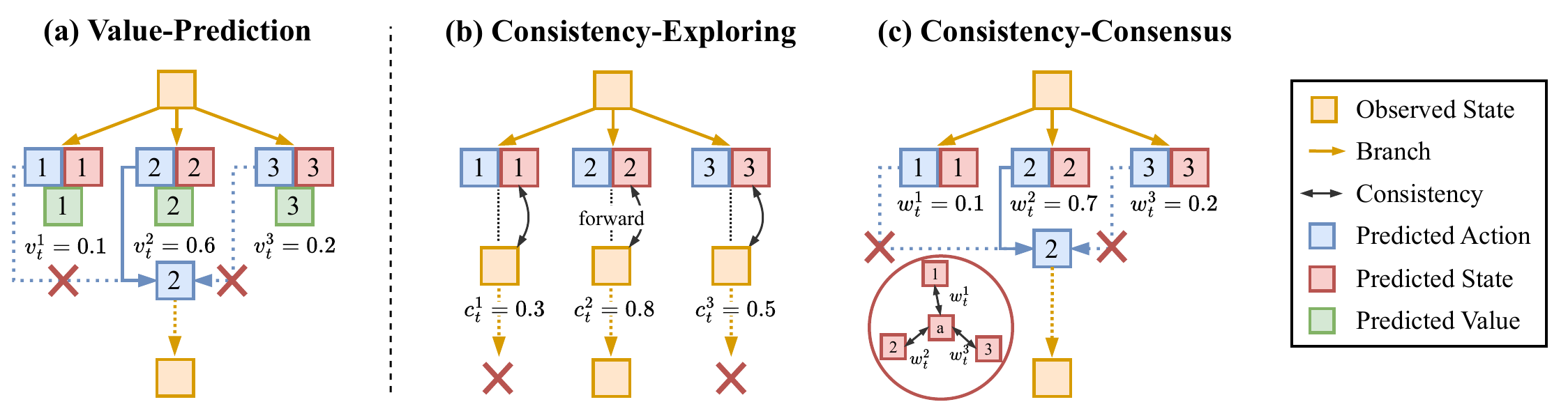}
    \caption{\textbf{Illustration of best-of-$N$ test-time selection strategies.} \textbf{(a) Value-Prediction}: each branch is scored by the model's predicted value function, and the branch with the highest score is selected. \textbf{(b) Consistency-Exploring}: each candidate branch is executed from the same initial state, and the branch with the highest consistency score is selected. \textbf{(c) Consistency-Consensus}: candidate future states are aggregated into a shared consensus future, and the branch whose predicted future is most consistent with this consensus is selected.}
    \label{fig:its_strategy}
    \vspace{-5pt}
\end{figure}

\subsection{Selection by Environment Exploring}

The most direct way to use consistency for test-time scaling is to explicitly evaluate each candidate branch after execution. Given the current state, the world action model samples $N$ candidate futures to obtain a sequence of actions and predicted future state $\{(\hat a_{t+1:t+\Delta}^{(i)}, \hat o_{t+\Delta}^{(i)})\}_{i=1}^N$. In the exploration setting, each branch is executed from the same initial state, producing an observed next observation $o_{t+\Delta}^{(i)}$. We then score each branch by its action-state consistency:
\begin{equation}
c_t^{(i)} = c_t\left(\hat o_{t+\Delta}^{(i)}, o_{t+\Delta}^{(i)}\right),
\end{equation}
and select the branch with the highest score (illustrated in~\cref{fig:its_strategy} (a)):
\begin{equation}
i_t^\star = \arg\max_{i \in \{1,\dots,N\}} c_t^{(i)}, 
\qquad 
a_t = \hat a_t^{(i_t^\star)}.
\end{equation}

This strategy estimates how well consistency can serve as a value signal. However, it assumes that the agent can reset to the same state and execute every candidate action, which is generally unavailable in real-world robotic settings. Therefore, we use this strategy primarily as an upper bound for measuring the potential of consistency-guided selection.

\subsection{Selection by Future Consensus}

Since environment rollout selection requires executing every candidate branch in the environment, it is impractical for real-world robotic deployment, where resetting to the same state is often unavailable or costly. We therefore seek a fully test-time strategy that approximates true action-state consistency without requiring environment resets or additional executions. Such a strategy should rely only on the model's sampled predictions, allowing all candidate branches to be evaluated in parallel before committing to a single action.

Our key assumption is that the consensus mean of predicted futures, $\bar o_{t+\Delta}$, provides a better approximation to the true future observation than any individual sample, as averaging multiple predictions reduces the impact of individual prediction errors. Under this assumption, a branch whose predicted future lies closer to the consensus mean is more likely to also lie closer to the true future, and therefore more likely to achieve higher true consistency. This motivates scoring each branch by its agreement with the consensus future as a deployable proxy for action-state consistency.
Formally, given $N$ predicted observations, we first average all to compute a consensus future :
\begin{equation}
\bar o_{t+\Delta} = \frac{1}{N} \sum_{j=1}^{N} \hat o_{t+\Delta}^{(j)}.
\end{equation}
Each branch is then scored by how consistent its predicted future is with the consensus:
\begin{equation}
c_t^{(i)} = c_t\!\left(\hat o_{t+\Delta}^{(i)}, \bar o_{t+\Delta}\right).
\end{equation}

We then select the branch with the highest consensus score via \textit{winner-takes-all}:
\begin{equation}
i_t^\star = \arg\max_{i \in \{1,\dots,N\}} c_t^{(i)},
\qquad
a_t = \hat a_t^{(i_t^\star)}.
\end{equation}

We prefer a winner-takes-all design over weighted action aggregation since robot actions do not always lie in a Euclidean space. Many control commands involve pose and rotation components associated with manifolds such as $SE(3)$ or $SO(3)$, where naive linear averaging may distort the underlying motion structure. Moreover, different branches can represent distinct motion hypotheses, and averaging them may blur their intent into an action unsupported by any predicted future. We provide further discussion in~\cref{sec:consensus_wta}.

\section{Experimental Evaluation}\label{sec:exp}

\paragraph{Setup.}
We conduct experiments with Cosmos-Policy~\citep{cosmospolicy} on RoboCasa~\citep{robocasa} and LingBot-VA~\citep{lingbotva} on RoboTwin 2.0~\citep{robotwin}, covering joint-prediction and inverse-dynamics WAM formulations, respectively. All experiments are run on 8 NVIDIA RTX 5090 GPUs. For test-time scaling (TTS), we use $N=8$ candidate rollouts by default. Following the parallel GPU setup of Cosmos-Policy~\citep{cosmospolicy}, candidate rollouts can be evaluated in parallel, so the additional wall-clock overhead is small in our implementation. Without parallelization, runtime scales approximately linearly with the number of candidates $N$. The overhead of computing selection weights and consistency scores is negligible, taking approximately $0.7$ ms for $N=8$.

\begin{wraptable}{r}{0.55\textwidth}
\vspace{-\baselineskip}
\centering
\caption{\textbf{Results on RoboCasa.} 
``$*$'' denotes our reimplementation; all other results are taken from~\citep{cosmospolicy}. 
\textit{TTS} indicates test-time scaling.}
\label{tab:robocasa_comp}
\resizebox{\linewidth}{!}{\begin{tabular}{lcc}
\toprule
Method & TTS & Average SR (\%) \\
\midrule
UVA~\cite{uva} & \ding{55} & 50.0 \\
DP-VLA~\cite{dpvla} & \ding{55} & 57.3 \\
UWM~\cite{uwm} & \ding{55} & 60.8 \\
$\pi_0$~\cite{pi0} & \ding{55} & 62.5 \\
GR00T-N1.5~\cite{gr00t} & \ding{55} & 64.1 \\
Video-Policy~\cite{videopolicy} & \ding{55} & 66.0 \\
FLARE~\cite{flare} & \ding{55} & 66.4 \\
\noalign{\smallskip}
\hdashline
\noalign{\smallskip}
Cosmos-Policy$^*$~\cite{cosmospolicy} & \ding{55} & 66.6 \\
+ Value-Prediction$^*$ & \checkmark & 67.4 \\
+ Consistency-Consensus (\textbf{ours}) & \checkmark & 67.3 \\
+ Consistency-Exploring (\textbf{ours}) & \checkmark & \textbf{68.0} \\
\bottomrule
\end{tabular}}
\end{wraptable}

\subsection{Results on RoboCasa}

We follow the evaluation protocol of Cosmos-Policy~\citep{cosmospolicy} on RoboCasa~\citep{robocasa}, which contains 24 kitchen manipulation tasks performed by a single Franka Emika Panda robot arm. For each task, we evaluate 50 trials and report the average success rate. \cref{tab:robocasa_comp} shows that the consistency-guided test-time selection improves the reimplemented Cosmos-Policy baseline from 66.6\% to 68.0\%. The improvement is modest but consistent with our central claim: action-state consistency can serve as a useful ranking signal even without an explicit value head. In particular, \textit{Consistency-Consensus} achieves a success rate of $67.3\%$, nearly matching value-based selection ($67.4\%$). \textit{Consistency-Exploring} obtains the highest success rate ($68.0\%$), as it directly executes and evaluates candidate branches in the environment, and therefore serves as an upper-bound strategy. Unlike value-based selection, \textit{Consistency-Consensus} does not rely on a learned value predictor, making it the more practical variant for deployment.

The qualitative result in~\cref{fig:compare_results}(a) provides a more direct view of this effect. The original rollout fails when the predicted future no longer matches the realized object transition during manipulation. By contrast, \textit{Consistency-Consensus} selects a branch whose predicted future is better supported by other sampled futures, yielding a higher consensus score, and leads to successful task completion. This suggests that consistency-guided selection can identify branches with more reliable action-conditioned consequences. More examples are provided in~\cref{sec:add_visual}.

\begin{wraptable}{r}{0.55\textwidth}
\vspace{-\baselineskip}
\centering
\caption{\textbf{Results on RoboTwin 2.0.}
``$*$'' denotes our reimplementation; all other results are taken from~\citep{lingbotva}.
\textit{TTS} indicates whether test-time scaling is applied.}
\label{tab:robotwin}
\resizebox{\linewidth}{!}{\begin{tabular}{lcc}
\toprule
Method & TTS & Average SR (\%) \\
\midrule
X-VLA~\cite{xvla} & \ding{55} & 72.9 \\
$\pi_0$~\cite{pi0} & \ding{55} & 65.9 \\
$\pi_{0.5}$~\cite{pi05} & \ding{55} & 82.7 \\
Motus~\cite{motus} & \ding{55} & 88.7 \\
\noalign{\smallskip}
\hdashline
\noalign{\smallskip}
$^*$LingBot-VA~\cite{lingbotva} & \ding{55} & 90.2 \\
+ Consistency-Consensus (\textbf{ours}) & \checkmark & \textbf{93.0} \\
\bottomrule
\end{tabular}}
\end{wraptable}

\subsection{Results on RoboTwin 2.0}
\begin{figure}[h]
    \centering
    \includegraphics[width=\linewidth]{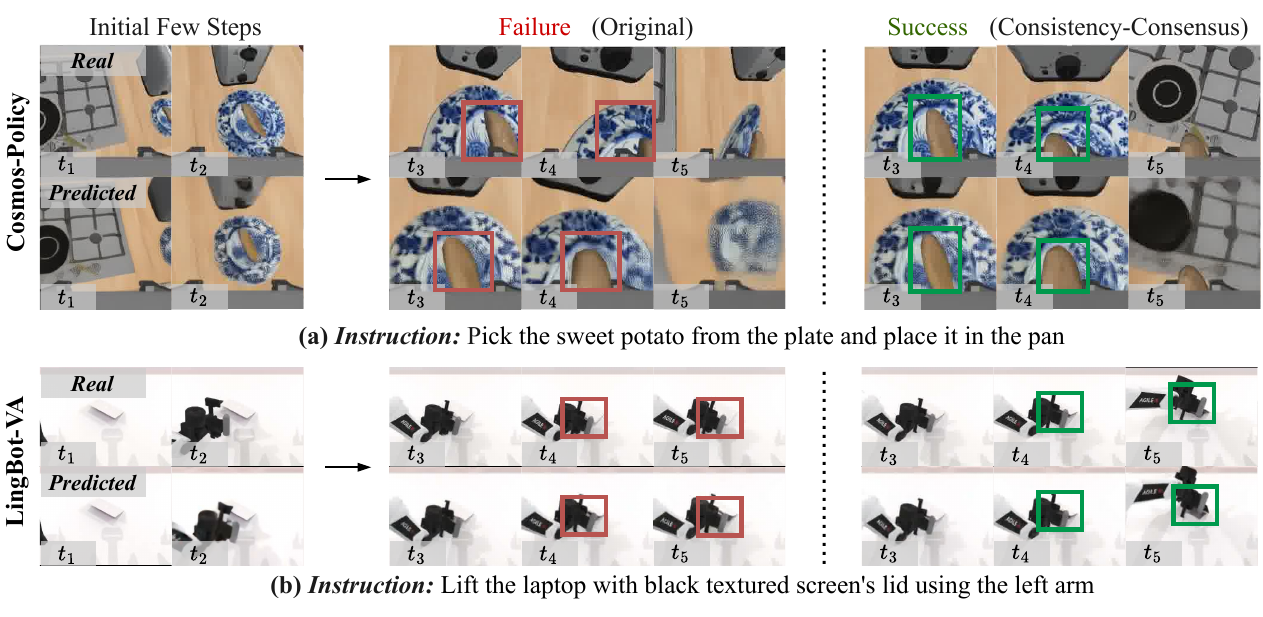}
    \caption{\textbf{Comparison between original and consistency-guided selection.}
We compare the original selection strategy with our consistency-guided selection across two \textbf{(a)} Cosmos-Policy and \textbf{(b)} LingBot-VA. Consistency-guided selection improves task execution by selecting action branches that better align predicted state transitions with the executed actions.}
    \label{fig:compare_results}
\end{figure}

We further evaluate consistency-guided selection on RoboTwin 2.0~\citep{robotwin}, following the setup of LingBot-VA~\citep{lingbotva}. RoboTwin 2.0 contains over 50 bimanual manipulation tasks requiring coordinated dual-arm control, providing a complementary testbed with a different embodiment and task structure from RoboCasa. We run 10 trials per task and report the average success rate under the \textit{Easy} setting, where each task uses a fixed initial configuration. The results in \cref{tab:robotwin} show that \textit{Consistency-Consensus} improves the reimplemented LingBot-VA baseline from 90.2\% to 93.0\%, supporting our findings on RoboCasa. 
Notably, since LingBot-VA follows an inverse-dynamics formulation, this result demonstrates that consistency-guided selection generalizes beyond joint-prediction WAMs. Furthermore, as RoboTwin 2.0 does not provide a state-saving API, \textit{Consistency-Exploring} cannot be evaluated; the improvement therefore comes entirely from the deployable \textit{Consistency-Consensus} variant. The qualitative comparison in~\cref{fig:compare_results}(b) further illustrates this effect. The original rollout fails to maintain the intended object-level progress, whereas \textit{Consistency-Consensus} selects a branch whose predicted transition is better aligned with the realized coordinated motion. Additional examples are provided in~\cref{sec:add_visual}.

\subsection{Scaling with Number of Candidates}

As discussed in~\cref{sec:rq2}, misaligned failures often arise when the policy enters a stalled regime and later predictions collapse toward low-motion backgrounds. In~\cref{sec:mitigate_bgc}, we provide additional evidence that consistency-guided selection can mitigate this behavior by selecting better early-stage actions, thereby helping the rollout avoid such collapsed regimes. This suggests that giving the selector more candidate futures at test time may further improve performance.

\begin{wrapfigure}[14]{r}{0.55\textwidth}
    \vspace{-\baselineskip}
    \centering
    \includegraphics[width=\linewidth]{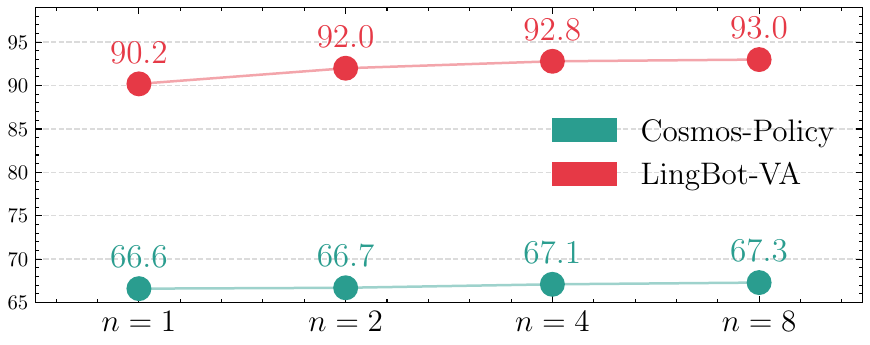}
    \vspace{-15pt}
    \caption{\textbf{Scaling consistency-guided selection with more candidate rollouts.} Increasing the number of sampled candidates $N$ improves the success rate for both models, showing that consistency provides an effective test-time scaling signal.}
\label{fig:scalable}
\end{wrapfigure}

We therefore examine whether consistency-guided selection benefits from additional test-time computation. Specifically, we vary the number of sampled candidates as $N \in \{1,2,4,8\}$ for both Cosmos-Policy and LingBot-VA, and select actions using the \textit{Consistency-Consensus} strategy. As shown in~\cref{fig:scalable}, increasing $N$ improves the success rate for both models. This trend indicates that a consistency-guided strategy provides a meaningful ranking signal over sampled candidate futures: as more candidates are available, the selector is more likely to find an action branch whose predicted future is well supported by the sampled consensus. Overall, these results show that our proposed consistency-guided strategy benefits from test-time scaling.

\section{Conclusion and Future Work}

This work studies action-state consistency as a reliability axis for world action models (WAMs). Across joint-prediction and inverse-dynamics formulations, we show that consistency is strongly associated with task success, follows similar success-failure trends as learned value prediction, and provides a value-free signal for test-time selection. We also identify background collapse as a key failure mode, where low-dynamics failed trajectories can become deceptively consistent. These results suggest that WAM reliability should be evaluated not only by task reward, visual prediction quality, or value estimates, but also by whether predicted futures remain faithful to the action-conditioned transitions they represent. More broadly, action-state consistency points toward consistency-aware world action modeling. Future WAMs should not only predict actions or futures, but also expose whether their imagined futures are dynamically compatible with the actions being considered. By making the reliability of imagined consequences explicit, consistency-aware WAMs offer a path toward robot policies that can reason not only about what actions to take, but also about whether their predicted outcomes can be trusted before acting in the world.
\clearpage

{
\small
\bibliographystyle{plainnat}
\bibliography{cite} 
}


\begin{appendices}
    \newpage
\appendix

{
\centering
\Large
\textbf{Appendix} \\
\vspace{1.0em}
}

\addcontentsline{toc}{section}{Appendix}

{
\hypersetup{linkbordercolor=white}
\startcontents[appendix]
\section*{Content}
\printcontents[appendix]{}{1}{}
}
\clearpage

\section{Implementation Details}\label{sec:impl}

We follow the official evaluation and inference setups of Cosmos-Policy\footnote{\url{https://github.com/nvlabs/cosmos-policy}} and LingBot-VA\footnote{\url{https://github.com/robbyant/lingbot-va}} for all experiments. Unless otherwise specified, we set the number of sampled candidate branches for test-time scaling to $N=8$.

\section{Related Work}

\paragraph{World Models.}
World models~\cite{world_model} learn predictive representations of environment dynamics and provide a powerful foundation for planning, control, and embodied decision making. By modeling how the world evolves under actions, they allow agents to reason about future consequences before execution, enabling reinforcement learning~\cite{safedreamer,rlvr}, imagination-based planning~\cite{dreamerv1,dreamerv2}, and data-efficient policy learning~\cite{dreamerv3}. Recent advances further extend world models to high-dimensional visual domains, where generative models can predict future observations and capture rich physical regularities from large-scale video data~\cite{genie,sora,worldsimbench,cosmos25,lingbot-world,leworld}. These capabilities make world models especially appealing for robotics, where agents must understand object motion, contact dynamics, and long-horizon scene evolution~\cite{daydreamer,mindv,roboticworldmodel,roboscape}. Building on this line of work, our paper studies world action models, which extend predictive world modeling by jointly modeling future observations and the actions that induce them.

\begin{wrapfigure}{r}{0.5\textwidth}
    \vspace{-\baselineskip}
    \centering
    \includegraphics[width=\linewidth]{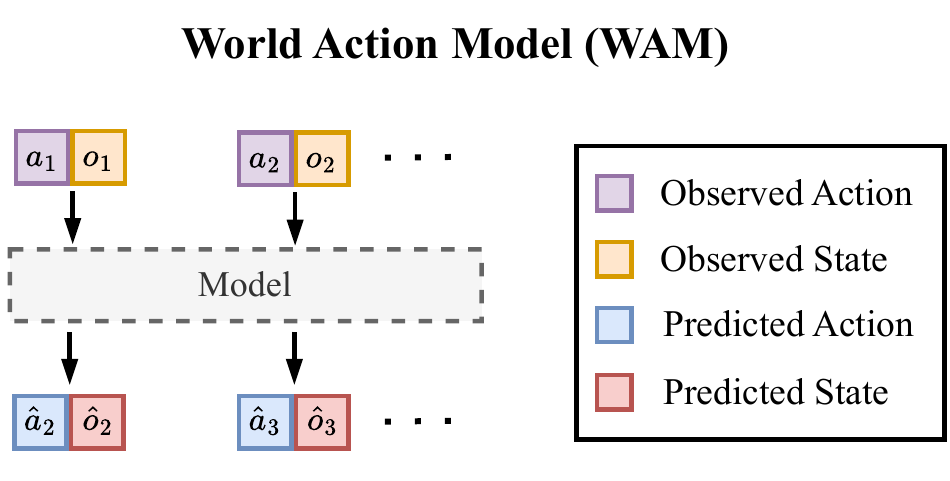}
    \caption{\textbf{Illustration of the World Action Model (WAM).} A WAM usually refers to a frame-prediction-based action model, where value prediction is optional for a best-of-N strategy.}\label{fig:wam_demo}
\end{wrapfigure}

\paragraph{World Action Models.}
Vision-language-action (VLA) models have emerged as a scalable paradigm for robot control by mapping visual observations and language instructions directly to executable actions~\cite{rt2,openvla,pi05,pi06,xvla,dpvla,alpamayo}. By leveraging large-scale vision-language pretraining and robot demonstration data, these models provide a general interface for instruction-conditioned manipulation across diverse tasks and environments. However, direct action prediction alone does not explicitly represent the future consequences of an action, making it difficult to inspect whether an action is physically plausible or aligned with the intended task outcome. World action models (WAMs) build on this progress by extending action prediction with future observation prediction~\cite{worldvla,motus,cosmospolicy,lingbotva,dreamzero,fastwam,gigaworld}. Instead of only asking what action should be taken, a WAM also asks what future state this action should induce. This joint prediction view makes the model's internal prediction more interpretable and actionable: the generated future can support planning, candidate selection, and self-evaluation. In this work, we study this joint structure through action-state consistency, measuring whether the predicted action and future state agree with the transition observed after execution, and how this agreement relates to downstream task outcomes.

\paragraph{Test-Time Scaling for Robot Control.}
Test-time scaling has recently emerged as an effective way to improve robot policies by allocating additional computation during deployment. Instead of only scaling training data, model size, or optimization steps for VLA models, different methods have been proposed to sample and refine multiple candidate actions with learned verifiers~\cite{rover}, scale instruction rephrasing and action verification~\cite{scalingverification}, select among candidates using internal confidence signals~\cite{verifierfree}, perform search over imagined futures~\cite{vlareasoner}, refine actions through latent iterative computation~\cite{recurrentvla}, or adapt policies online from environment feedback~\cite{evolvevla}. For WAMs, Cosmos-Policy~\cite{cosmospolicy} proposes using predicted future states and a value head to perform best-of-$n$ selection over generated candidates. These methods show that additional inference computation can improve robustness and task success, but they generally require some scoring signal, such as an external verifier, a learned value function, task-specific feedback, or model-specific confidence. In contrast, we aim to use the inner structure of WAMs together with environment observations to derive a direct test-time scoring signal. Since not all WAMs include a value head, we show that action-state consistency can serve as a model-agnostic signal that can be directly applied across different WAMs.

\section{Per-Task Analysis}\label{sec:across_tasks}

\subsection{Consistency Score}\label{sec:consistency_across_tasks}

\begin{figure}[!h]
    \centering
    \begin{subfigure}{\linewidth}
        \centering
        \includegraphics[width=\linewidth]{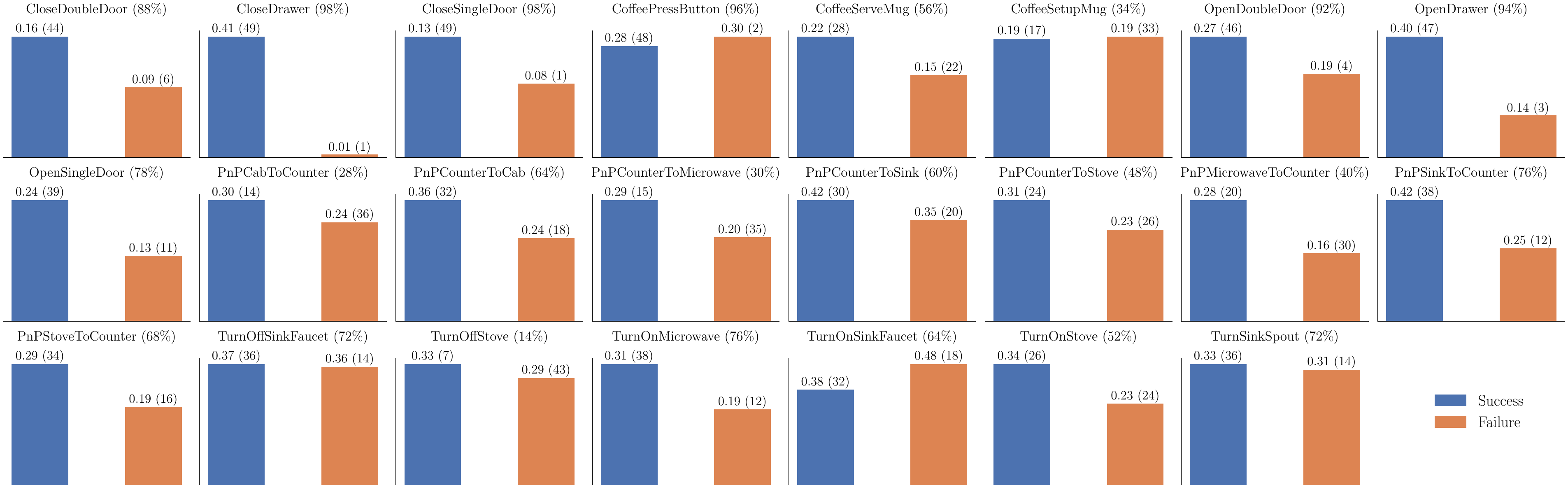}
        \caption{\textbf{Cosmos-Policy.} Per-task consistency scores on RoboCasa, comparing successful and failed episodes. We show 23 out of 24 tasks.}\label{fig:cosmos_error_task}
    \end{subfigure}
    
    \vspace{0.5cm}

    \begin{subfigure}{\linewidth}
        \centering
        \includegraphics[width=\linewidth]{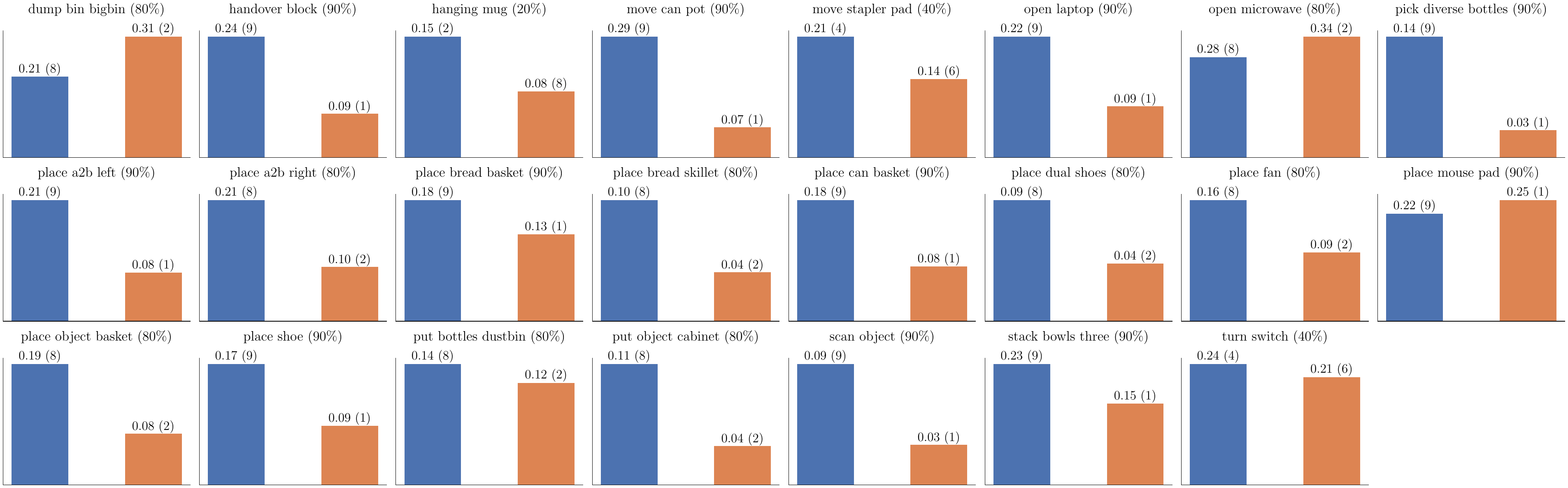}
        \caption{\textbf{LingBot-VA.} Per-task consistency scores on RoboTwin 2.0, comparing successful and failed episodes. We show 23 out of 50 tasks.}\label{fig:lingbot_error_task}
    \end{subfigure}
    
    \caption{\textbf{Consistency across tasks.} We report per-task consistency scores for successful and failed episodes. Across most tasks, successful episodes exhibit a higher consistency score than failures. Tasks with full success are excluded for clearer comparison.}
    \label{fig:consistency_across_task}
\end{figure}

In addition to the overall analysis in~\cref{sec:rq1}, we provide a per-task breakdown of consistency scores in Figure~\ref{fig:consistency_across_task}, comparing successful and failed episodes within each task.

Across both Cosmos-Policy (RoboCasa) and LingBot-VA (RoboTwin 2.0), we observe that in most tasks, successful episodes exhibit a higher consistency score than failed ones. This trend is especially pronounced in some tasks, where the gap between success and failure is large (\eg, \texttt{CloseDrawer}, \texttt{pick diverse bottles}. This suggests that future prediction is closely tied to results.

However, the strength of this relationship varies across tasks. In a few cases, failed episodes exhibit comparable or lower error than successful ones (\eg, \texttt{CoffeePressButton} and \texttt{place mouse pad}). As discussed in~\cref{sec:rq2}, these exceptions are largely explained by \emph{background collapse}, where stalled low-motion episodes become easier to predict and can therefore receive favorable consistency scores. Overall, the per-task results further support consistency as a meaningful indicator of task progress, while motivating a deeper study of the conditions under which it becomes less reliable.

\subsection{ROC Curve}\label{sec:auc_across_task}

\begin{figure}[h]
    \centering
    
    \begin{subfigure}{0.49\linewidth}
        \centering
        \textbf{{\small (a) Cosmos-Policy}}
        \includegraphics[width=\linewidth]{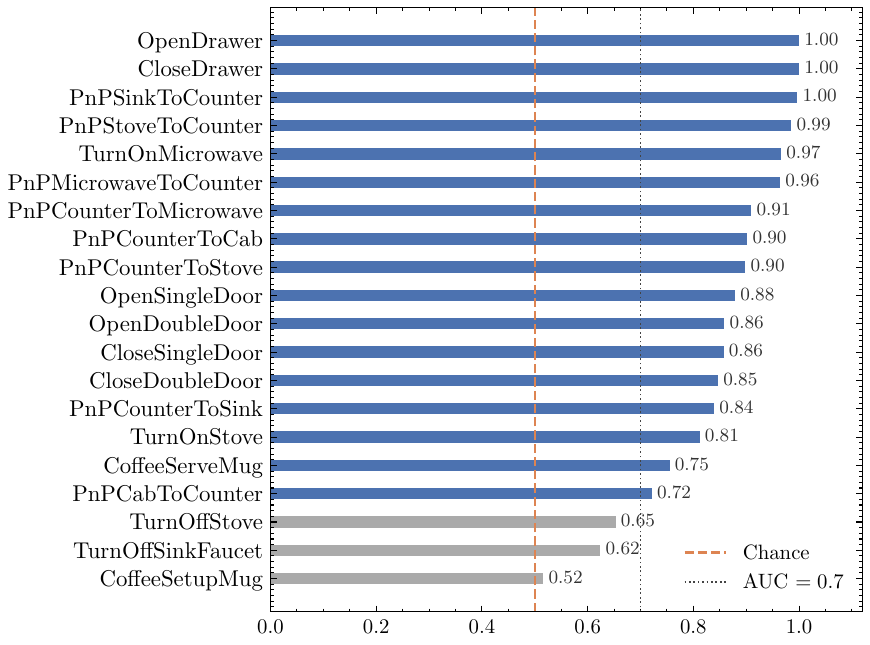}
    \end{subfigure}
    \hfill
    \begin{subfigure}{0.49\linewidth}
        \centering
        \textbf{{\small (b) LingBot-VA}}
        \includegraphics[width=\linewidth]{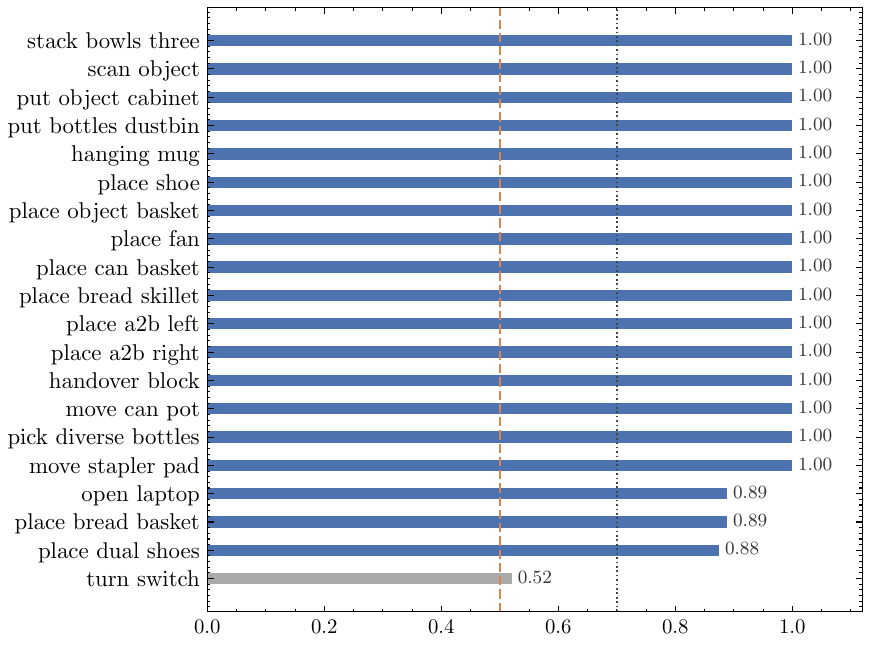}
    \end{subfigure}
    
    \caption{\textbf{Per-task ROC curves.} We include only tasks where successful episodes exhibit higher consistency than failed ones, to better analyze the predictive signal of consistency.}
    \label{fig:auc_across_task}
\end{figure}

We further report per-task ROC curves in~\cref{fig:auc_across_task} to analyze how the predictive power of consistency varies across tasks. We observe strong performance across many tasks, and the resulting AUC is often high and can approach $1.0$, indicating near-perfect separability. This is particularly evident in many manipulation tasks across both datasets (AUC $>0.9$).

For tasks with smaller consistency gaps, the corresponding AUC values are more moderate, reflecting weaker but still informative separation. In particular, tasks with AUC values near chance level (AUC $\approx 0.5$) typically exhibit almost identical consistency scores between successful and failed episodes, making the two outcomes difficult to distinguish using consistency alone. As discussed in~\cref{sec:rq2}, this weak separation can often be traced to low-motion or background-collapse regimes, where failed episodes remain visually static and are therefore easy to predict.

Overall, these results show that consistency provides a strong predictive signal for task success in many settings, while its effectiveness naturally depends on the extent to which future-state alignment reflects meaningful task progress.

\section{Relationship between Motion Change and Consistency Score}
\label{sec:relationship}

\begin{wraptable}{r}{0.42\linewidth}
    \vspace{-10pt}
    \centering
    \small
    \caption{\textbf{Correlation between latent change and consistency.}}
    \label{tab:motion_consistency_corr}
    \vspace{-5pt}
    \begin{tabular}{lcc}
        \toprule
        Model & Pearson & Spearman \\
        \midrule
        Cosmos-Policy & $-0.47$ & $-0.46$ \\
        LingBot-VA    & $-0.35$ & $-0.30$ \\
        \bottomrule
    \end{tabular}
    \vspace{-10pt}
\end{wraptable}

\begin{figure}[h]
    \vspace{-5pt}
    \centering

    \begin{subfigure}{0.49\linewidth}
        \centering
        \textbf{{\small \textcolor[HTML]{212121}{(a) Cosmos-Policy}}}
        \includegraphics[width=\linewidth]{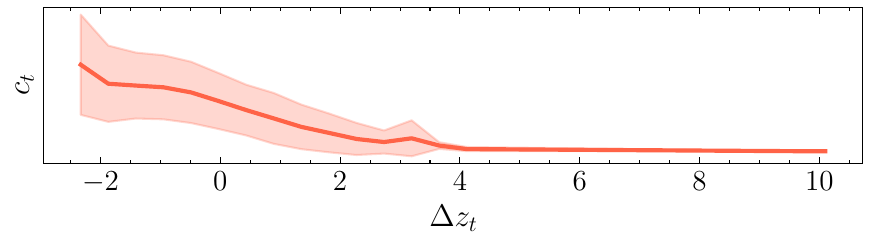}
    \end{subfigure}
    \hfill
    \begin{subfigure}{0.49\linewidth}
        \centering
        \textbf{{\small \textcolor[HTML]{212121}{(b) LingBot-VA}}}
        \includegraphics[width=\linewidth]{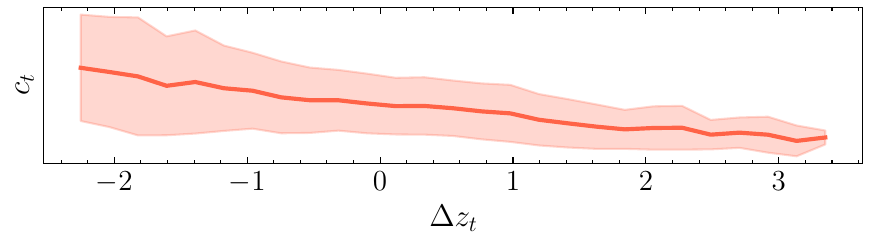}
    \end{subfigure}

    \vspace{4pt}

    \begin{subfigure}{0.49\linewidth}
        \centering
        \includegraphics[width=\linewidth]{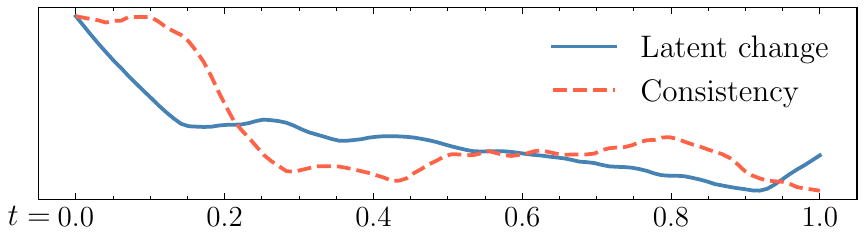}
    \end{subfigure}
    \hfill
    \begin{subfigure}{0.49\linewidth}
        \centering
        \includegraphics[width=\linewidth]{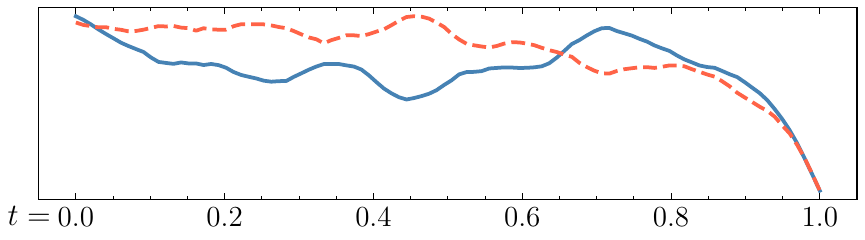}
    \end{subfigure}

    \caption{\textbf{Relationship between latent change and consistency.}
    \textbf{(top)} Plots showing that larger latent changes generally correspond to lower consistency scores across both models. 
    \textbf{(bottom)} Temporal traces showing that latent change and consistency often evolve in opposite directions over time. 
    Together, these results suggest that low-dynamics transitions are easier to predict and therefore tend to receive higher consistency scores.}
    \label{fig:motion_consistency_combined}
\end{figure}

To further analyze how transition dynamics affect consistency, we examine the relationship between latent transition magnitude and consistency score. Before visualization, both quantities are normalized into $z$-scores within each model to make their scales comparable. As shown in~\cref{fig:motion_consistency_combined} and~\cref{tab:motion_consistency_corr}, the two variables exhibit a clear negative correlation: larger latent changes tend to correspond to lower consistency scores, while smaller latent changes tend to receive higher consistency scores. Beyond this relationship, the two signals also tend to evolve in opposite directions over time: when $\Delta z_t$ rises, indicating stronger motion or larger state change, the consistency score typically falls; when $\Delta z_t$ decreases, consistency tends to increase. These results support our observation that consistency is affected by scene transition difficulty, with low-motion transitions being easier to predict consistently.

\section{Algorithm of Consistency-Guided Test-Time Scaling Strategies}
\label{sec:algo}

We now describe two algorithms for using action-state consistency as a test-time scaling signal introduced in~\cref{sec:consistency_guided}. At each decision step, the world action model samples $N$ candidate branches, where each branch predicts both an action sequence and its resulting future observation. The first strategy, \emph{Consistency-Exploring}, directly evaluates each branch through environment execution and selects the branch with the highest action-state consistency. The second strategy, \emph{Consistency-Consensus}, avoids environment execution by scoring each branch according to its agreement with the consensus among predicted futures. The full procedures are summarized in Algorithms~\ref{alg:consistency_explore} and~\ref{alg:consistency_consensus}.

\paragraph{Notation.}
At timestep $t$, let the model produce $N$ candidate branches
$\{(\hat a_{t+1:t+\Delta}^{(i)}, \hat o_{t+\Delta}^{(i)})\}_{i=1}^{N}$,
where $\hat a_{t+1:t+\Delta}^{(i)}$ is the predicted action and $\hat o_{t+\Delta}^{(i)}$ is the predicted next observation of branch $i$. Let $c_t(\cdot,\cdot)$ denote the consistency score in~\cref{eq:con_score}.

\begin{algorithm}[h]
\caption{Consistency-Exploring}
\label{alg:consistency_explore}
    \begin{algorithmic}[1]
    \Require Current observation $o_t$, number of branches $N$
    \State Sample candidate branches $\{(\hat a_{t+1:t+\Delta}^{(i)}, \hat o_{t+\Delta}^{(i)})\}_{i=1}^{N}$
    \For{$i=1$ to $N$}
        \State Reset environment to state $o_t$
        \State Execute candidate action $\hat a_{t+1:t+\Delta}^{(i)}$
        \State Observe next state $o_{t+\Delta}^{(i)}$
        \State Compute score:
        $c_t^{(i)} \gets c_t(\hat o_{t+\Delta}^{(i)}, o_{t+\Delta}^{(i)})$
    \EndFor
    \State Select best branch:
    $i^\star \gets \arg\max_i c_t^{(i)}$
    \State Reset environment to state $o_t$
    \State Execute $\hat a_{t+\Delta}^{(i^\star)}$
    \end{algorithmic}
\end{algorithm}

\begin{algorithm}[h]
\caption{Consistency-Consensus}
\label{alg:consistency_consensus}
    \begin{algorithmic}[1]
    \Require Current observation $o_t$, number of branches $N$
    \State Sample candidate branches $\{(\hat a_{t+1:t+\Delta}^{(i)}, \hat o_{t+\Delta}^{(i)})\}_{i=1}^{N}$
    \State Compute consensus future:
    $\bar o_{t+\Delta} \gets \frac{1}{N}\sum_{i=1}^{N}\hat o_{t+\Delta}^{(i)}$
    \For{$i=1$ to $N$}
        \State Compute agreement score:
        $c_t^{(i)} \gets c_t(\hat o_{t+\Delta}^{(i)}, \bar o_{t+\Delta})$
    \EndFor
    \For{$i=1$ to $N$}
        \State Compute weight:
        $w_t^{(i)} \gets \frac{\exp(c_t^{(i)}/\tau)}{\sum_j \exp(c_t^{(j)}/\tau)}$
    \EndFor
    \State Select best branch:
    $i^\star \gets \arg\max_i w_t^{(i)}$
    \State Execute $\hat a_{t+\Delta}^{(i^\star)}$
    \end{algorithmic}
\end{algorithm}

\section{Why Does Consensus Use Winner-Takes-All Selection?}
\label{sec:consensus_wta}

A natural alternative to branch selection is to compute a weighted average of predicted actions using consensus scores. However, in our final design, we adopt winner-takes-all selection and directly execute the highest-scoring branch.

\begin{wraptable}[16]{r}{0.3\textwidth}
    \vspace{-12pt}
    \centering
    \caption{\textbf{Winner-takes-all (WTA) vs. weighted consensus action aggregation.} Directly averaging actions degrades success rate, while selecting the highest-scoring branch performs better.}
    \label{tab:wta}
    \resizebox{\linewidth}{!}{\begin{tabular}{lc}
    \toprule
    Method & Average SR (\%) \\
    \midrule
    \multicolumn{2}{c}{\textbf{Cosmos-Policy}} \\
    \noalign{\smallskip}
    \hdashline
    \noalign{\smallskip}
    + Weighted & 64.9 \\
    + WTA & \textbf{67.3} \\
    \midrule
    \multicolumn{2}{c}{\textbf{LingBot-VA}} \\
    \noalign{\smallskip}
    \hdashline
    \noalign{\smallskip}
    + Weighted & 86.4 \\
    + WTA &  \textbf{93.0} \\
    \bottomrule
    \end{tabular}}
\end{wraptable}

The main reason is that robot actions are not always represented in a globally linear space. Many control commands involve end-effector pose, orientation, or pose increments, where rotational components are naturally associated with nonlinear manifolds such as $SO(3)$ or $SE(3)$. In such cases, a naive Euclidean average of multiple candidate actions may fail to preserve a physically meaningful motion command and can degrade performance.

Even when the averaged action remains numerically valid, it can still be behaviorally inconsistent. Different branches often correspond to distinct motion hypotheses, such as approaching from different directions or selecting different grasp trajectories. Averaging these hypotheses may blur their intent and produce an action that is not supported by any predicted future.

To verify this effect, we implement a weighted consensus baseline that converts consistency scores into normalized weights through a temperature-scaled softmax, and then averages the predicted actions:
\begin{equation}
a_t = \sum_{i=1}^{N} w_t^{(i)} \hat a_t^{(i)}, \quad
w_t^{(i)} =
\frac{\exp(c_t^{(i)} / \tau)}
{\sum_{j=1}^{N} \exp(c_t^{(j)} / \tau)},
\quad \tau > 0.
\end{equation}

We compare this design against winner-takes-all selection in~\cref{tab:wta}. The results show that direct action averaging consistently underperforms selecting the highest-scoring branch, indicating that preserving the coherence of a single motion hypothesis is more effective than mixing multiple candidate actions.
\section{Mitigating Background Collapse}\label{sec:mitigate_bgc}

\begin{wrapfigure}{r}{0.55\textwidth}
    \centering
    \vspace{-\baselineskip}
    \includegraphics[width=\linewidth]{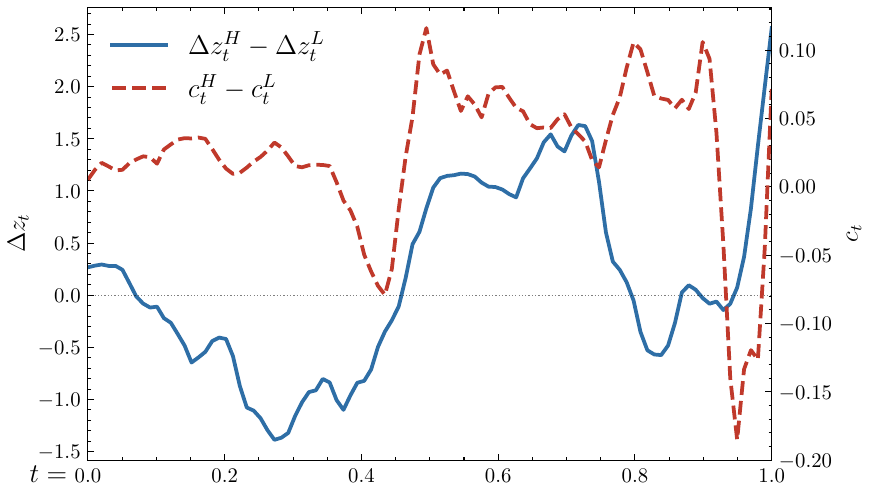}
    \caption{\textbf{Mitigating background collapse through improved consistency.} We compare original rollouts with lower success rate (\textbf{L}) against consistency-guided rollouts with higher success rate (\textbf{H}).}
    \label{fig:collapse_mitigation}
    \vspace{-10pt}
\end{wrapfigure}

\cref{sec:rq2} identifies background collapse as a failure mode in which predicted trajectories become nearly static after the policy enters a stalled regime. If this failure is partly caused by poor early action choices, then improving test-time action selection should reduce the chance of entering such low-dynamics states. We test this hypothesis by comparing lower-success vanilla rollouts with higher-success rollouts selected by consistency-guided exploration. As shown in~\cref{fig:collapse_mitigation}, we plot the difference between the higher-success consistency-guided rollout ($\textbf{H}$) and the lower-success vanilla rollout ($\textbf{L}$), denoted as \textbf{H}$-$\textbf{L}.
Positive values indicate that the consistency-selected rollout exhibits a larger latent transition magnitude or higher consistency than the vanilla rollout at the same timestep.  As shown in the figure, these differences are mostly positive in the later stages of execution, where background collapse is more likely to occur. This suggests that consistency-guided selection helps preserve task-relevant dynamics, rather than allowing the rollout to collapse into static predictions. These results support our interpretation that better early-stage action selection can reduce the likelihood of entering background-collapse regimes.
\clearpage
\section{Limitations and Failure Analysis}\label{sec:limitation}

\begin{wrapfigure}[11]{r}{0.55\textwidth}
    \vspace{-\baselineskip}
    \centering
    \includegraphics[width=\linewidth]{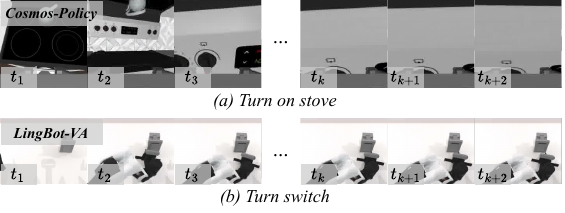}
    \caption{\textbf{Failure cases of consistency-guided selection having background collapse.}}
    \label{fig:limitation}
\end{wrapfigure}

Although consistency provides a value-free test-time signal for ranking candidate branches, it does not directly correct failures in the underlying world action model. For instance, background collapse can arise when the model predicts futures that preserve scene appearance while failing to capture task-relevant object interactions. Consistency-guided selection can mitigate this issue by choosing better candidate branches, but it cannot guarantee that collapsed futures will not occur, as shown in~\cref{fig:limitation}. Since this failure is largely rooted in training-time representation and dynamics learning, fully removing background collapse likely requires training-time interventions, such as stronger consistency supervision or data collection that emphasizes task retry under the failure scenarios.
\section{Additional Examples of Background Collapse}\label{sec:add_bg_cases}

\begin{figure}[h]
    \centering
    \includegraphics[width=\linewidth]{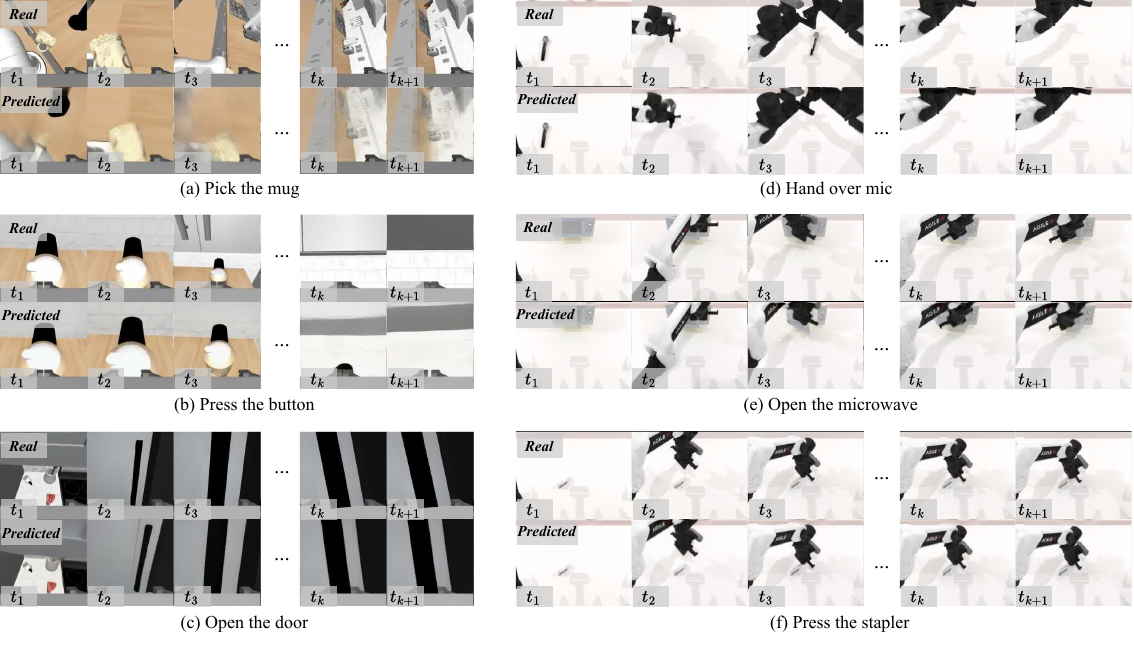}
    \caption{\textbf{Additional Examples of Background Collapse.} Qualitative examples from RoboCasa (left columns) and RoboTwin 2.0 (right columns). In consecutive predicted frames, such as $t_k$ and $t_{k+1}$, the scene often becomes nearly static or preserves only background content while task-relevant object motion disappears.}
    \label{fig:bg_collapse_more}
\end{figure}

We provide additional qualitative examples of background collapse in \cref{fig:bg_collapse_more} across both RoboCasa and RoboTwin 2.0. Background collapse refers to a degenerate prediction mode in which the generated future frames become nearly static, exhibit minimal scene change, or repeatedly preserve only the background while ignoring task-relevant object dynamics. This often occurs when the policy prematurely treats the task as complete and stops producing meaningful actions (\eg, case \textbf{(b)} and \textbf{(e)}), or when the manipulated object leaves the camera view and enters configurations that are poorly represented in the training distribution (\eg, case \textbf{(d)}).

Such behavior can be identified most clearly by comparing consecutive predicted frames, such as $t_k$ and $t_{k+1}$, where little motion is observed despite the task not being successfully finished. In these cases, reduced temporal variation can artificially increase consistency scores, since static predictions are easier to match across steps.

We emphasize that background collapse is not exclusive to consistency-misaligned tasks and can also appear in consistency-aligned settings. However, tasks where consistency fails to align with success outcomes tend to exhibit this phenomenon more frequently, making collapse a common source of misleadingly favorable consistency scores.

\section{Additional Visualizations}
\label{sec:add_visual}

\begin{figure}[h]
    \centering
    \includegraphics[width=\linewidth]{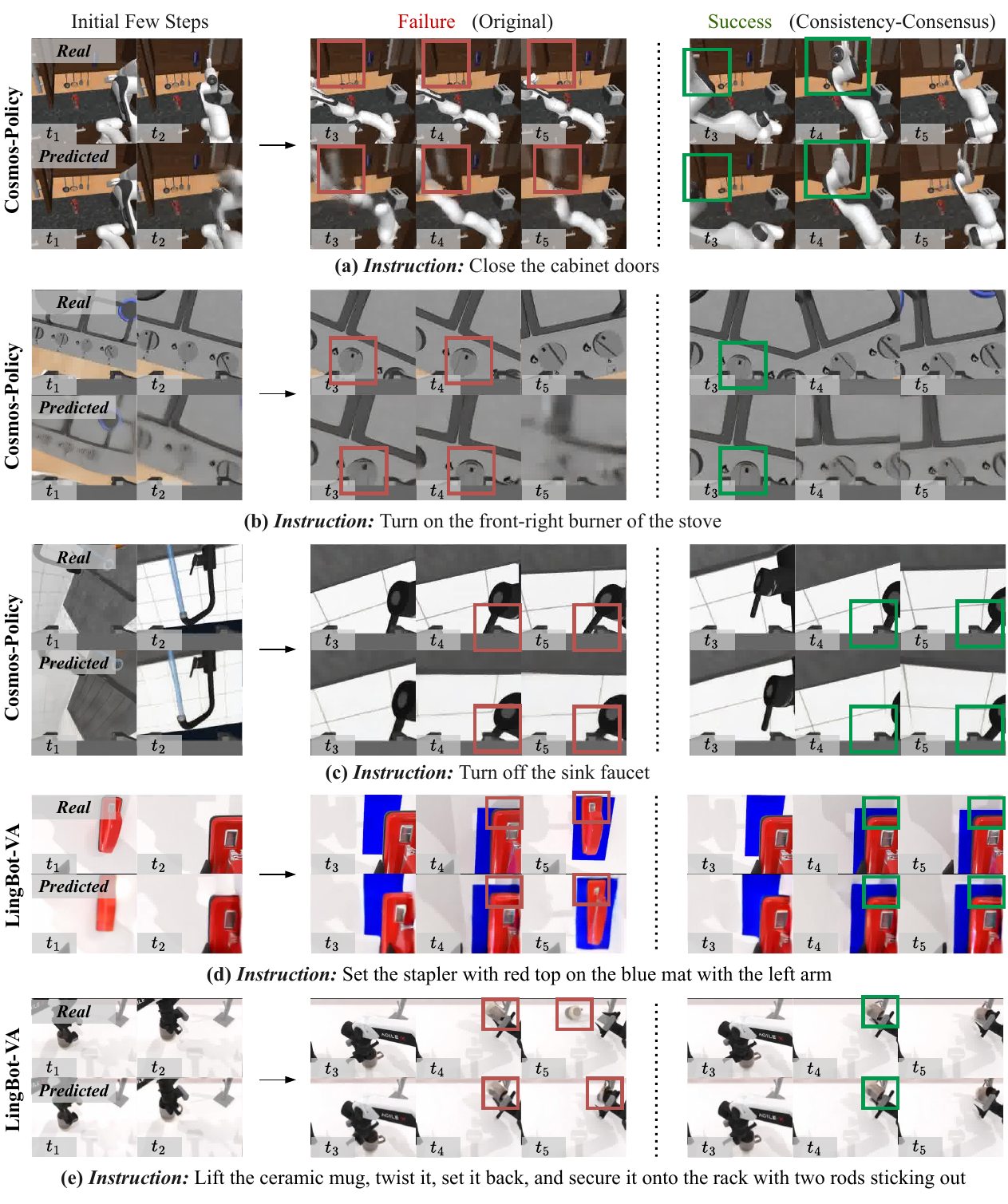}
    \caption{\textbf{Additional qualitative examples of success and failure results with consistency.} The visualizations show that discrepancies between predicted and realized futures often correspond to task-relevant errors, such as incorrect object state, failed grasping, blocked motion, or misaligned placement. Cases~(a) and~(d) are shown from different robot views to better visualize the interaction.}
    \label{fig:add_compare}
\end{figure}

We provide additional qualitative visualizations in~\cref{fig:add_compare} to illustrate how inconsistency between predicted and realized futures can affect task execution. These examples show that consistency errors often correspond to meaningful failure modes. For instance, the robot may predict that a door is closed when it is actually open, causing it to execute an inappropriate action and become stuck~\textbf{(a)}. In other cases, the model may incorrectly predict that an object will be grasped~\textbf{(b)}, that the arm can move without obstruction~\textbf{(c)}, that a stamp will remain within the paper boundary~\textbf{(d)}, or that a mug handle is properly aligned for hanging~\textbf{(e)}. These discrepancies between predicted consequences and actual outcomes explain why consistency can serve as a useful signal for selecting among candidate actions. For cases~\textbf{(a)} and~\textbf{(d)}, we adopt a different robot view for better visualization.

\section{Additional Statements}

\subsection{Use of LLMs}\label{sec:use_of_llm}

We use large language models to support the writing process, specifically for paraphrasing and improving the flow and clarity of the text. The LLM is used as an editing aid to refine wording and smooth paragraph transitions, while the core ideas, analysis, and final content remained our own.

\subsection{Broader Impacts}\label{sec:broad_impact}

This work studies the properties of world action models for robotics and is primarily foundational rather than a deployed robotic system. Nevertheless, improved world action models could eventually support more capable autonomous robots, which may raise safety, security, and societal concerns if used in high-stakes or uncontrolled environments. Potential negative impacts include unsafe behavior caused by model errors, misuse for harmful automation or surveillance, and uneven performance across user groups or environments due to biased training or evaluation data. We mitigate these concerns by focusing on analysis rather than deployment, documenting limitations, and encouraging careful evaluation, monitoring, and human oversight before any real-world robotic use.
\end{appendices}


\end{document}